\documentclass{article}
\usepackage{graphicx}
\begin{document}
\title{{\bf Human-Computer Conversation}}
\author{{\bf Yorick Wilks and Roberta Catizone}\\
University of Sheffield, Department of Computer Science \\
Regent Court, 211 Portobello Street \\
Sheffield, S1 4DP, UK \\
{\tt \{yorick, roberta\}@dcs.shef.ac.uk}}
\date{}
\maketitle

\begin{abstract}
The article surveys a little of the history of the technology, sets out
the main current theoretical approaches in brief, and discusses the
on-going opposition between theoretical and empirical approaches.  It
illustrates the situation with some discussion of CONVERSE, a system
that won the Loebner prize in 1997 and which displays features of both
approaches.
\end{abstract}

\section{Introduction}

Practical and theoretical investigations into the nature and structure
of human dialogue have been a topic of research in artificial
intelligence and the more human areas of linguistics for decades: there
has been much interesting work but no definitive or uncontroversial
findings.  The best performance overall in HMC (Human-machine
conversation) has almost certainly been Colby's PARRY program
\cite{colby71} since its release on the (then ARPA) net around 1973.
It was robust, never broke down, always had something to say and,
because it was intended to model paranoid behaviour, its zanier
misunderstandings could always be taken as further evidence of mental
disturbance, rather than the processing failures they were.  Colby
actually carried out a version of the Turing test \cite{turing50}
--usually taken to be the test of whether users can tell a human
from a machine in a HMC environment--by getting psychiatrists to
compare, blind, PARRY utterances with those of real paranoids and they
were indeed unable to distinguish them.  Indistinguishability results are never statisticfally watertight, but it was, nonetheless, a very striking demonstration and much longer ago than many now realise.

\section{History}

In theoretical terms, modelling conversation or dialogue comes within
the division of linguistics that Morris \cite{morris64} called 
pragmatics--that
which deals with the relationship of individual agents to symbols and
the world--as opposed to syntax, that deals with the relationship of
symbols to each other, and semantics, that deals with the relationship
of symbols to the world.  HMC is a pragmatic study, but its history has
involved many attempts to see it in syntactic and semantic terms as
well, since the more tractable areas of research and pragmatics was
often thought too hard.

If we look back at the HMC systems of the Seventies we see a clear
division of approach that persists to this day: on the one hand there
were theoretically-motivated models in the artificial intelligence
tradition that emphasized reasoning and a deep understanding of
language based on knowledge of the world about which the conversation
took place.  The most famous was Winograd's program \cite{win72} that
discussed a set of blocks on a table top, and more realistic settings
were Grosz's \cite{grosz77} system that acted as a robot discussion of how to
assemble a water pump.\\

{\parindent 0pt
If you said 
}\\

{\it ``Attach the platform to the pump"} \\

{\parindent 0pt
it could reply 
} \\

{\it ``Where are the bolts?"} \\

{\parindent 0pt
although no bolts had been mentioned because it knew from its knowledge
structure that bolts were needed for assemblies.  The most advanced
system of its generation was the Toronto train timetable system \cite{allen78}
that would reply to: \\
}

{\it ``The 6.15 to Vancouver?"} \\

{\parindent 0pt
with \\
}

{\it ``Platform 6"} \\

{\parindent 0pt
In some sense, it knew what was wanted by a person saying the first
utterance, which is just a noun phrase and not even a  question.  None
of these systems had very good performance and their vocabularies were
a few dozen words at most; they were never tested by being opened up to
a public who could talk to them in an unconstrained way--their systems
of syntactic and semantic analysis were far too fragile for that--if
one word was differently placed they might understand nothing at all.
}

A quite different kind of system was Colby's PARRY, mentioned
above--and which was never as famous as its far far weaker contemporary
ELIZA \cite{weiz67}.  PARRY knew little of the world, had no syntax analysis
and just worked by a large set (about 6000) of patterns with which it
matched any input.  It appeared to be a paranoid patient in a Veterans'
hospital and in reply to: \\

{\it ``Have you been hospitalized before"} \\

{\parindent 0pt
it might say \\
}

{\it ``No, this is the first time"} \\

{\parindent 0pt
The essence of PARRY was not in individual answers but in the fact that
it could keep up a conversation like these for dozens of turns and
appeared to have things it wanted to say, particularly stored
resentments about the Mafia and bookies at the race track.  It appeared
to have a personality and many users (and it had thousands) refused to
believe it was a computer at all.
}

This dichotomy between theory-driven models and performance driven
models persists to this day.  One could classify the different
approaches to HMC roughly as follows.

\section{The current state of play}

One could argue that Human Machine Conversation (HMC) is now in a
position like that of machine translation fifteen years ago: it is a
real technology, coming into being in spite of scepticism, but with a
huge gulf between busy practitioners getting on with the job, often
within companies, and, on the other side, the researchers, in
linguistics, artificial intelligence (AI) or whatever, whose papers
fill the conferences.  The rise of empirical linguistics has largely
closed this gulf for machine translation and related arts, but not as
yet, for HMC.  First, let us set out historical trends in HMC, for if
the HMC world is really as disparate as what follows it is hardly
surprising there is so little consensus on how to progress.

\begin{enumerate}

\item Dialogue analysis based on models of individual agents' beliefs
and knowledge structures, usually presented within an AI-Derived theory
of plans, inference and possibly speech/dialogue acts and truth
maintenance, many using a ``space" metaphor to represent individuals
(e.g. Allen \cite{allen82}, Traum \cite{traum96}, Kobsa \cite{kobsa86}, 
Ballim and Wilks \cite{ballim91}).

\item AI-derived models of dialogue based on more linguistic notions
and not primarily based on models of individuals; the representation is
often in terms of partitioned semantic nets to represent domains but
uses concepts of focus, failure, repair etc.  (e.g. Grosz and Sidner
\cite{grosz89}, Webber \cite{webber79}).

\item AI-derived models based largely on transitions in domain scripts,
driving top-down inference but augmented by local inference rules
representing quasi-plan etc., no models of individuals (e.g. Schank
\cite{schank77}).

\item Sociology/Ethnomethodology tradition of descriptive conversational
analysis, usually based on local transition analysis types in actual
dialogues, analysed non-statistically (e.g. Schegloff \cite{scheg}).

\item Local AI theories of discourse, usually without models of
individuals or domains, but with a taxonomy of speech/dialogue acts and
inference rules applied bottom-up to utterances (Charniak \cite{char93}, 
Bunt \cite{bunt89}, Carletta \cite{carl92}).

\item Transition network models of general discourse moves--apparently
global but domain-independent scripts but largely dependent on local
alternatives--a dialogue version of ``text grammar" and in some ways a
normalization of (4) above (e.g. Whittaker and Stanton \cite{whit88}).

\item Empirical analysis of dialogue corpora to produce statistical
measure of dialogue turn transitions based on a taxonomy of dialogue
actsùthe first empirical pragmatics? In some ways, the provision of
evidence for forms of (6) and so (4).  (e.g. Maier and Reithinger
\cite{rei95,alex97}).Pattern-matching approaches to bottom-up dialogue
analysis, providing input to some higher representational form and
rejecting the possibility of effective dialogue grammar (Colby 
\cite{colby73}, ATIS \cite{pall92} and CONVERSE \cite{levy97}).

\item Full treatment of empirical dialogue analysis, derived 
automatically from corpora, transducing from utterances to some form rich
enough to support dialogue acts, whether in terms of some conventional
grammatical representation or a fusion of taggers, lexical strings and
pattern matcher outputs.

\end{enumerate}

{\parindent 0pt
The themes and approaches in this list are probably not wholly
independent and may not be exhaustive.  Notice that, thirty years after
PARRY, no real form of (9) exists, and the corpora from which it might
be done (for English at least) have only very recently come into
existence.  An interesting question right now, is whether (9) can be
done in a principled way, as an alternative to PARRY-like systems built
up over long periods by hand, or many of the other types of systems
above with trivial vocabularies and virtually no functionality.  This
was exactly the opposition in machine translation for many years: with
SYSTRAN's large hand-crafted functionality contrasted with a host of
theoretical, published, acclaimed but non-functional systems.  In
Machine Translation, that opposition began to collapse with the arrival
of IBM's statistical MT system about 1990.  The possibility of a
meaningful empirical pragmatics could do the same for HMC.
}

One additional point should be made here: we have said nothing of
computer recognition (and production) of speech in dialogue systems.
Speech research has pursued its own agenda, separate from written text,
and all the above systems communicated via screen typing.  The chief
speech problem was always decoding the signals into words, rather than
the content of dialogue as we have described it above--researchers
tended to assume that speech could be solved separately and then a
dialogue model of one of the following types just bolted on, as it
were.  This agenda for research has had obvious defects, especially in
that speech phenomena like pauses, stress, pitch etc.  convey meaning
as well--but basically there has been agreement on all sides until now
to separate out the speech and language issues so as to progress.

\section{The Loebner Prize}

The Loebner Prize Medal is awarded annually to the designer of the
computer system that best succeeds in passing a variant of the Turing
Test, in which human judges communicate with a workstation and try to
decide which of the systems in the competition is a program and which a
person.  The winning program is the one the judges are least able to
distinguish from the human interlocutors taking part.  Complex
competition rules control typing speeds and so on, so that the machine
entries do not give themselves away by typing too fast.The competition
is overseen by the ACM, the main US organization for computer
professionals, and for the last two years there has been no domain
restriction on what can be talked about: programs entering must in
principle be prepared to talk about anything at all.  CONVERSE (the
1997 winner) had strong views on the lesbian couple Bill Clinton had
welcomed to the White House the night before the competition, and of
course on Clinton himself.  It narrowly beat out the 1998 winner, an
Australian program that claimed to be a 14-year old girl marooned on 
a desert island and appealing for help over the World Wide Web.

Competitions have included American, Canadian and Australian programs
and it is the first time it has been won by a non-US team.  David Levy,
Director of Intelligent Research, claimed that in twenty years people
will be falling in love with these programs, and they are certainly
more stimulating than Tamagotchi pets, and of a far higher standard
than the Chatterbots currently available on the Web.  The example below
simply repeats exactly the kind of limited paraphrase, masquerading as
chat, that was to be found in Schank's programs in the early
1970's.  More details on past and future competitions and full
transcripts of the 1997 Loebner competition can be found on the Web
site: http://acm.org/~loebner/loebner-prize.htmlx.  A sample of
CONVERSE's 1997 performance is at the end of this article.

\section{CONVERSE}

The CONVERSE program was not intended to be based on any scientific
research but on hunches about how to do it, while taking advantage of
some recent methodological shifts in computational linguistics towards
empiricism and the use of real data.  The main hunch was derived
directly from PARRY's impressiveness when compared with its passive
contemporaries like ELIZA. (Weizenbaum, 1976) PARRY had something to
say, just as people do, and did not simply react to what you said to
it.  It could be said to rest on the hunch that a sufficient condition
for humanness in conversation may be what Searle \cite{searle83} 
calls intentionality: the apparent desire to act and affect surroundings 
through the conversation, which is a strong version of what we are
calling ``having something to say", since a computer program without
prostheses can only create such effects through speech acts and not
real acts on physical objects.  The extension of this hunch as far as
Turing test situations are concerned -- i.e.  fooling people that the
system is human--is that if the computer can get to say enough, to keep
control of the conversation, as it were, through being interesting or
demanding so that the human plays along, then there is correspondingly
less opportunity for the human interlocutor to ask questions or get
responses to an unconstrained range of utterances that will show up the
system for what it is.  Naturally enough, this hunch must be tempered
in practice since a system that will not listen at all, and which will
not be diverted from its script no matter what is said, is again
inevitably shown up.  The hunch is simply that, and translatable as: be
as active and controlling in the conversation as
RACTER, PARRY's only real rival (as regards interestingness) over the
last 30 years, worked on the principle of being so interesting and zany
that many humans did not want to interrupt it so as to intrude new
topics or demands of their own.  Others were less charmed of course,
but it was one very effective strategy for operating this key hunch,
and one not involving clinical madness, as PARRY did.

{\parindent 0pt
The original features of CONVERSE are as follows: 
}

\begin{enumerate}

\item top down control of conversation by means of a range of scripts,
plus an active bottom up module seeking to answer questions etc. using
a set of data bases on individuals.

\item control and interaction of these features in (1) by means of a
weighting system between modules that could be set so as to increase
the likelihood of one or another of these modules "gaining control" at
any given point in the conversation.

\item the use of large scale linguistic data bases such as thesaurus
networks giving conceptual connectivity--for dealing with synonyms--and
large proper name inventories (Collins' dictionary proper names in our
case) that allowed CONVERSE to appear to know about a large range of
people and things not otherwise in the scripts, the data bases, or the
semantic nets, though this dictionary information was formally mapped
to the structures of the semantic network and the databases.

\item a commercial and very impressive text parser, based on trained
corpus statistics.  This however had been trained for prose rather than
dialogue which meant that much of its output had to be modified by us
before being used.  We also made use of large scale patterns of
dialogue use derived from an extensive corpus of British dialogue that
has recently been made available.  
\end{enumerate}

{\parindent 0pt
The last takes advantage of recent trends in natural language
processing: the use of very large resources in language processing and
intermediate results obtained from such resources, like the dialogue
patterns.  It meant that CONVERSE was actually far larger than any
previous Loebner entry, and that much of our effort had gone into
making such resources rapidly available in a PC environment.  So,
although not based on specific research, CONVERSE was making far more
use of the tools and methods of current language processing research
than most such systems. Its slogan at this level was ``big data, small 
program" which is much more the current trend in language processing
and artificial intelligence generally than the opposite slogan, one
which had ruled for decades and seen all such simulations as forms of
complex reasoning, rather than the assembly of a vast array of cases
and data.  CONVERSE, although, it has some of the spirit of PARRY, does
in fact have data bases and learns and stores facts, which PARRY never
did, and will allow us in the future to expand its explicit reasoning
capacity.  The weighting system would in principle allow great
flexibility in the system and could be trained, as Connectionist and
neural network systems are trained, to give the best value of the
weightings in terms of actual performance.  We will continue to 
investigate this, and whether weightings in fact provide a good model
of conversation--as opposed to purely deterministic systems that, say,
always answer a question in the same way when it is posed.  In the end,
as so often, this may turn out to be a question of the application
desired: a computer companion might be more functionally appropriate if
weighted, since we seem to like our companions, spouses and pets to be
a little unpredictable, even fractious.  On the other hand, a computer
model functioning as a counselor or advisor in a heath care situation,
advising on the risks of a certain operation or test, might well be
more deterministic, always answering a question and always telling all
it knew about a subject when asked.
}

\subsection*{
The CONVERSE Personality}

The apparent character of CONVERSE is Catherine, a 26 year-old female
editor for a magazine like Vanity Fair, who was born in the UK, but
currently lives in New York.  The contents of Catherine's character are
stored in a database of features and values, known as the Person
database (PDB).  The kinds of things that we store about Catherine are
the details of her physical appearance, her birthday, astrological
sign, some of her likes and dislikes, whether she has a boyfriend,
where she works, etc.  For the most part, things in the PDB are all
related to facts about Catherine.  We can also store information about
other people in the PDB, in particular people that are related to
Catherine in some way: her mother, father, friend, boss, etc.  Scripts
are the driving force of the program and whenever possible, we aim to
keep control of the conversation, by posing a question at the end of a
system utterance. The scripts cover a range of 80 topics, but this can 
be easily extended (within the limits of the hardware) with a graphical
script editing interface.  Currently, some of the topics covered are
crime, racism, religion, the Simpsons, mobile phones, abortion, travel,
food and violence.  These are only differences of topic not of genre or
style.  The method for acquiring the scripts is done in a two stage
process: first, a script writer sketches out the script on paper and
secondly, the scripts are entered into the system via a semi-automatic
script editor.  The script editor establishes the flow of control
through the script based on the user's responses to each script
utterance.

Amongst the recreational applications which are foreseen for CONVERSE,
foremost is the idea of using it as a virtual friend.  CONVERSE's
Person Data Base can be augmented with data for different
personalities, enabling the user to talk, on one day, to a virtual
elderly English gentleman and on another occasion to a virtual
20-year-old punk music fan from the Bronx.  Eventually the user should
be able to describe the personality with whom he wishes to talk, and
there would then be a module to create a suitable Person Data Base
conforming to the user's specification.

\subsection*{Extending CONVERSE}

CONVERSE was based on quite simple intuitions: that conversational
skill is a compromise between two tendencies.  First, there is the
active, top-down, intentional driver with something to say (the feature
that Colby's PARRY (Colby 1971) had and Weizenbaum's ELIZA (Weizenbaum
1967) so clearly lacked).

Secondly, there was the passive, bottom-up, listener aspect which meant
understanding what was said to it and react appropriately, by answering
questions or even changing the topic.  This, as all researchers know,
is much harder because it requires understanding.  Humans who lack it
are conspicuously bad conversationalists but this is normally
attributed to not listening rather than not understanding what is
said.  We could call the simple CONVERSE architecture (Figure 1)
Pushmepullyou (after Dr Doolittle) to convey the tension between the
two elements.

\begin{figure*} 
\centerline{\includegraphics[width=10cm,height=14cm]{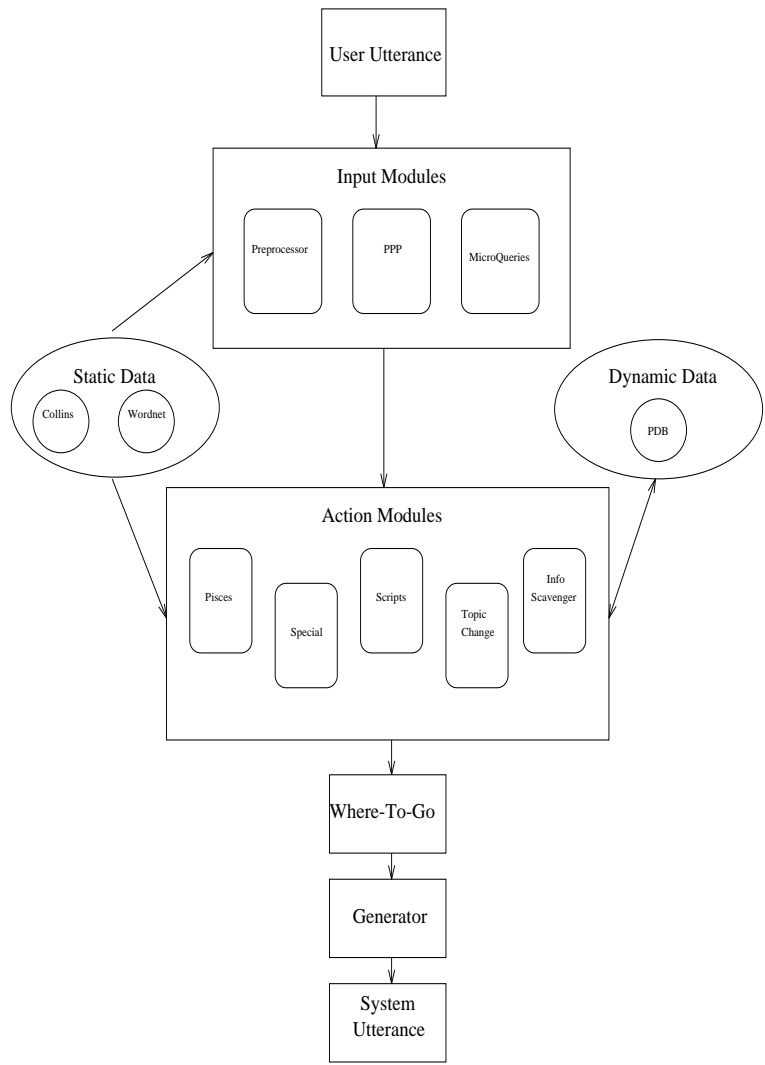}}
\caption{The CONVERSE Architecture}
\end{figure*}

{\parindent 0pt
What we are now engaged in is an attempt to move both the push and pull
sides to a higher level.  For the latter we hope to use a model of
individual agents beliefs and intentions we have worked on (in
Sheffield and New Mexico) for some years, called ViewGen \cite{lee96},
it is a well-developed method (with several
iterations of Prolog programs) of creating and manipulating the
beliefs, goals etc. of individual spaces for inference we call
environments, all controlled by an overall default process called
ascription that propagates beliefs etc. from space to space with
minimum effort so the system can model the states of other agents it is
communicating with.
}

It is based on the general assumption that a communicating agent must
model the internal states of its interlocutor agents as best it can,
not just by storing their features, like age and size, but their own
states.  This system is much stronger than the rather elementary Person
Data Base in CONVERSE but we intend to strengthen that with aspects of
ViewGen so as to increase its functionality substantially.

The more immediately challenging move is to replace the push-me or
parsing side of CONVERSE, which was based on a statistical parser of
general English prose with a shell of what we called microqueries which
adapted it to dialogue from normal prose.

We are currently working on a robust parser of English conversation,
which is to say, a transducer from sentences to a set of dialogue acts
plausible for any domain.  Now is a perfect moment to do this since
corpora of English dialogue (like the British National Corpus or BNC)
have now become available so that this task can be seen as an extension
of contemporary empirical computational linguistics into the field of
pragmatics itself, the last bastion.  The chief difficulty, suffered by
all researchers in this tradition, is that although the BNC is
available for unsupervised training, there is very little dialogue
corpus (Apart from Edinburgh's MAPTASK \cite{ander92}, Rochester's TRAINS 
and a corpus at VerbMobil) which is marked with a
set of dialogue acts for supervised training and evaluation.  There is
currently an international initiative (DAMSL) \cite{core} in
this area but as yet few signs of progress.  The next key step in
empirical linguistics, one not yet achieved anywhere, will be such a
robust model of English dialogue structure, probably using machine
learning methods, a task so far only attempted for speech interfaces in
very narrow domains.

\section{Conclusion}

It is worth emphasizing again how surprising it is that only recently,
and within the Loebner competition, have levels of performance in HMC
got back to the PARRY levels of nearly thirty years ago, in spite of
all the theoretical activity since then--it has not led to much robust
conversational performance.  Even now if you check out the Chatterbots
on the web they are of astonishingly low quality, far, far below that
of the Loebner winners.  A recent one, Peedy, chosen at random, told a
user that: \\

{\parindent 0pt
Just type in a sentence like: 
}

{\it John gave Mary a book because it was her birthday} \\

{\parindent 0pt
and then you can ask the corresponding questions: 
}

{\it Did you give Mary a book}
 
{\it What did John give Mary} 

{\it Who gave Mary a book} \\

{\parindent 0pt
and so on.
}

Those who look back at Schank's inference system of 1970 will see that
the example comes straight from there which is yet further proof, if it
were needed, of how little movement there has been in some theoretical
and empirical areas.  In spite of its brusque approach to a complex
technology, only the Loebner competition has been able to refocus R\&D
on robust and realistic performance, avoiding both the traps of theory
without performance, and the performance trivialities like Chatterbot
just quoted.

\newpage
\section{Appendix: CONVERSE Output from \\ Loebner Competition}

{\parindent 0pt
{\tt PROGRAM[10:18:04]Hi, my name is Catherine, what's yours?

JUDGE04[10:18:32]My name is Elliott.

PROGRAM[10:18:34]Well, it's nice to meet you, Elliott.

PROGRAM[10:18:48]Did you see that story on CNN last night about the lesbian

PROGRAM[10:19:04]couple who came out at a White House party on Sunday?

JUDGE04[10:19:55]Yes, I did.  I think it may all be a publicity stunt for Ellen.

PROGRAM[10:20:01]Ellen Degeneres was one of them - she was kissing her lover

PROGRAM[10:20:11]in front of Bill Clinton. Do you think he was very

PROGRAM[10:20:25]embarrassed?

JUDGE04[10:21:00]I think he want the gay vote. He will do anything to get it and keep it.

JUDGE04[10:21:15]He does not have a conservative base.

PROGRAM[10:21:21]

PROGRAM[10:21:21]Are you still a Friend of Bill Clinton?

JUDGE04[10:22:20]I am not sure if he has any real friends. He has achieved alot and has

JUDGE04[10:22:31]left alot of people behind.

PROGRAM[10:22:35]You really don't like the guy--well, at least that's

PROGRAM[10:22:50]definite--I thought most people were just indifferent or

PROGRAM[10:23:02]numb by now.

JUDGE04[10:23:34]I think we could have a more honest and dynamic leader.  One able to

JUDGE04[10:23:55]face the real dangers facing the country and address them with

JUDGE04[10:24:26]courage and intelligence. Rather then listening to public opinion polls.

PROGRAM[10:24:30]I guess that means you really think there's serious behind

PROGRAM[10:24:56]Whitewater, the Special prosecutor and all that stuff?

JUDGE04[10:25:58]We have two dead people, Brown and guy who commited suicide, we have lost

JUDGE04[10:26:36]papers. And we have Clintons people willing to tell all.  I think that

JUDGE04[10:26:55]Hillary will be found guilty of something.

PROGRAM[10:26:58]But how do you actually feel about him, now in the number

PROGRAM[10:27:09]two term?

PROGRAM[10:27:15].

JUDGE04[10:27:44]He is too smart to get caught. He deserved to named hotel manager of the

JUDGE04[10:27:47]year.

PROGRAM[10:27:49]OK, but what happens now--does it all drift on, with

PROGRAM[10:28:01]everyone waiting to see how much his friends have said about

PROGRAM[10:28:17]him under threat of lots and lots of jail-or will something

PROGRAM[10:28:32]dramatic actually happen?}
}

\end{document}